\documentclass{article}
\usepackage{spconf,amsmath,graphicx}

\usepackage{multirow}
\usepackage{subcaption}

\usepackage{enumitem}
\setlist{nosep, leftmargin=14pt}

\usepackage{mwe} 

\title{Brain Surface Reconstruction from MRI Images Based on Segmentation Networks Applying Signed Distance Maps}
\name{Heng Fang$^{\star \dagger}$ \qquad Xi Yang$^{\star}$ \qquad Taichi Kin$^{\star}$ \qquad Takeo Igarashi$^{\star}$}
\address{$^{\star}$ The University of Tokyo, Tokyo, Japan \\
     $^{\dagger}$ KTH Royal Institute of Technology, Stockholm, Sweden}
\begin{document}
%
\maketitle
\begin{abstract}
Whole-brain surface extraction is an essential topic in medical imaging systems as it provides neurosurgeons with a broader view of surgical planning and abnormality detection. To solve the problem confronted in current deep learning skull stripping methods lacking prior shape information, we propose a new network architecture that incorporates knowledge of signed distance fields and introduce an additional Laplacian loss to ensure that the prediction results retain shape information. We validated our newly proposed method by conducting experiments on our brain magnetic resonance imaging dataset (111 patients). The evaluation results demonstrate that our approach achieves comparable dice scores and also reduces the Hausdorff distance and average symmetric surface distance, thus producing more stable and smooth brain isosurfaces.
\end{abstract}
\section{Introduction}
\label{sec:intro}
Skull stripping is a significant task in medical image processing, as it not only protects patients' brains during intraoperative navigation and radiotherapy but also provides visualized information to conduct surgical planning and clinical-oriented diagnosis. Furthermore, skull stripping is a preprocessed step used to identify abnormalities, such as tumors, lesions, and cancerous cells; therefore, the quality of extracted brains would greatly affect the accuracy of abnormal tissue detection \cite{fatima2020state}. However, skull stripping is also a challenging task, as the brain is considered to be the most complex organ in the human body. To achieve great performance in the skull stripping process, it is necessary to obtain a stable and smooth brain isosurface.

Currently, brain surface extraction in clinical imaging is mostly conducted by experienced radiologists, outlining the brain boundary in each magnetic resonance imaging (MRI) slice. Obviously, this manual routine is prone to error and time consuming; therefore, semi-automatic end-to-end methods have emerged to improve both efficiency and fault tolerance.
A considerable number of conventional skull stripping algorithms have been proposed and widely used in brain MRI analysis, such as the Brain Extraction Tool (BET), Brain Surface Extractor (BSE), FreeSurfer, and the Hybrid Watershed Algorithm (HWA) \cite{rehman2020conventional}. However, conventional skull stripping algorithms such as deformable-surface-based methods require additional steps and extra computation cost \cite{wang2011robust,segonne2004hybrid}. As a result, recent studies have concentrated on applying deep learning methods to the skull stripping problem. Chen et al. proposed VoxResNet---a voxel-wise neural network architecture---to predict the label of each voxel and carefully designed the layers and connections to incorporate low-level images and high-level semantics \cite{chen2018voxresnet}. Afterwards, Hwang et al. validated that 3D U-Net, the most famous encoder--decoder semantics segmentation network in the field of medical imaging, can achieve state-of-the-art performance in the problem of skull stripping \cite{hwang20193d}. However, Geirhos et al. used quantitative experiments to show that convolution neural networks place more emphasis on texture information rather than shape information \cite{geirhos2018imagenet}, while brain surfaces have relatively stable positions and shapes. The aforementioned skull stripping methods predict whole brains without prior learning shape information, which is not suitable for brain surface reconstruction.

To better exploit shape information of brain isosurfaces, we introduce the learning mechanism of signed distance fields (SDFs), a concept from geometry modeling, into the skull stripping backbone network for the first time. The magnitude of SDFs implies the distance between the point and the closest boundary of the whole brain, and the sign of SDFs provides information on whether the point is inside the brain. Several works validated that compared to segmentation maps, SDFs contain more information, especially global shape information \cite{ma2020distance,kervadec2019boundary,karimi2019reducing,navarro2019shape,wang2020deep}. However, SDFs have not applied for brain surface reconstruction in literature. Besides, we introduce an additional Laplacian loss that exploits the property of SDFs to obtain a smoother, more continuous brain surface.

The main contributions of our work are as follows: (1) We applied 2D U-Net and 3D U-Net segmentation backbone networks to our brain MRI dataset and show that 2D U-Net performs better than 3D U-Net in the task of large organ surface extraction. (2) We added a regression head to the 2D U-Net backbone network to learn the information of SDFs and train the head jointly with the segmentation head. As a result, our new pipeline achieves better results in the evaluation metrics of the Hausdorff distance (HD) and average symmetric surface distance (ASSD). (3) We also introduced the Laplacian loss that uses the property of SDFs as an additional term in the loss function of the regression head and demonstrated that the new loss could help to reduce HD and obtain smoother brain isosurfaces in three-dimensional(3D) scenarios.

\section{Methods}
\label{sec: methods}
\subsection{Overview}
\begin{figure}[htbp!]
    \centering
    \includegraphics[width=1.0\linewidth]{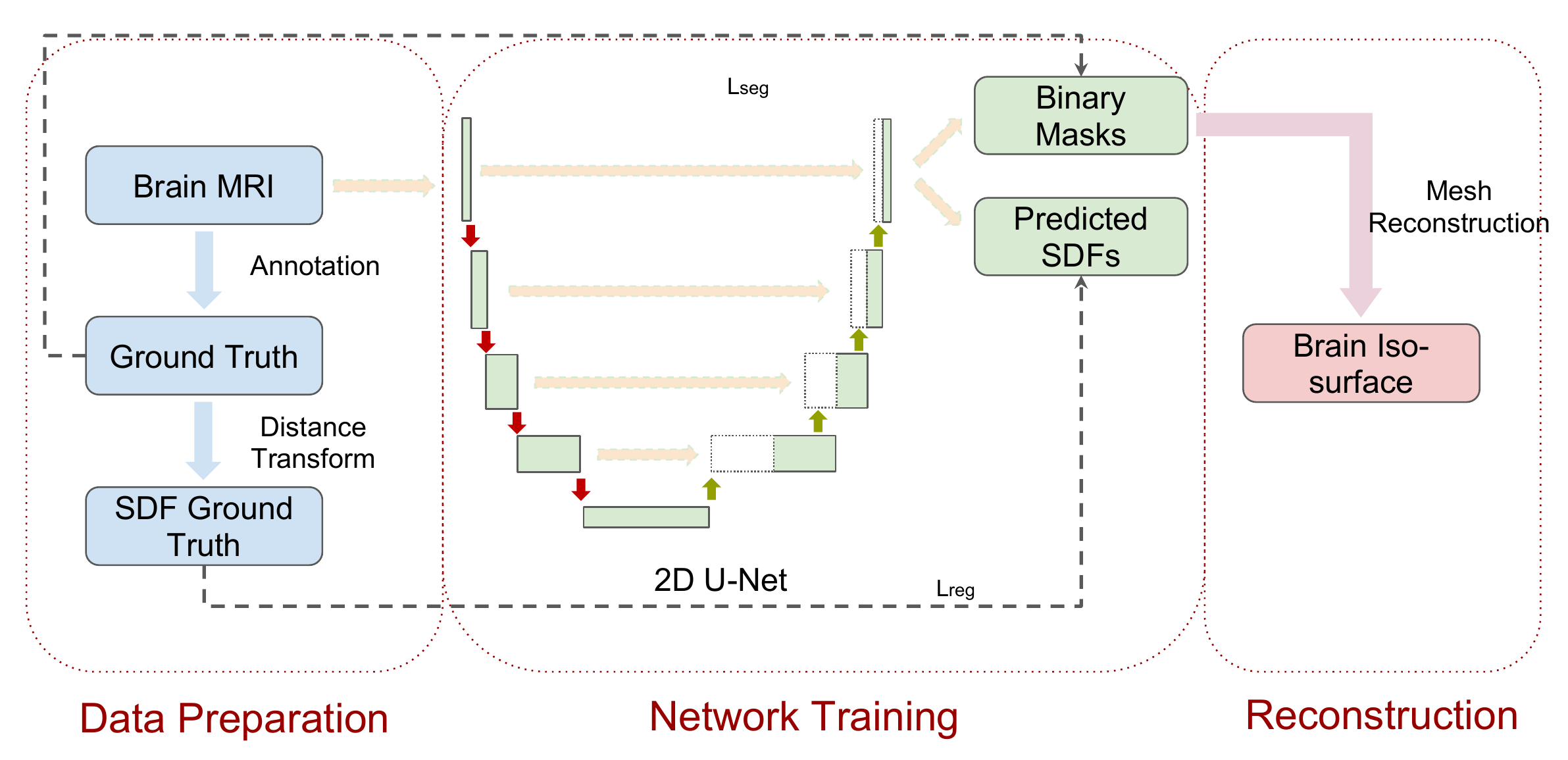}
    \caption{Pipeline of our method.}
    \label{fig:pipeline}
\end{figure}

We propose to add signed distance maps to the segmentation network
for whole brain surface reconstruction. We first perform a two-dimensional distance transformation on the ground truth separately in terms of each slice, and then train a 2D-U-Net-based convolution neural network with two heads to predict binary segmentation maps and scaled regression distance maps. Finally, we exploit the marching cube algorithms to generate 3D surfaces from segmentation maps for each patient. The pipeline is shown in Fig. \ref{fig:pipeline}.

\subsection{Signed Distance Fields Transform}
Given the ground truth of a brain surface defined for each brain slice $\Omega$ and a point $x$, the mapping between $\Omega$ and the SDFs is defined as 
\begin{equation*}
  SDF(x) =
\begin{cases} 
+d(x, \partial \Omega),  & \mbox{if }x \in \Omega_{out} \\
-d(x, \partial \Omega),  & \mbox{if }x \in \Omega_{in}
\end{cases}
\end{equation*}
where $\partial \Omega$ represents the boundary of each brain slice, $d(x, \partial \Omega)$ refers to the closest Euclidean distance between the point $x$ and the boundary (infimum), and the sign implies whether the point is inside or outside of the brain slice. In addition, the zero-level set of points can be resolved as the isosurface of the brain. In our method, we adopted 2D pixel distance fields, which represent the distance to the nearest pixel boundary between inside and outside, to calculate SDFs. 

\subsection{Network Architecture}
The network architecture is based on 2D U-Net, which employs an analysis path and a synthesis path to learn local and global features separately \cite{ronneberger2015u}. To predict the segmentation maps and regression distance maps simultaneously, we added a head to the existing architecture and trained the parameters with the ground truth and SDFs at the same time. Furthermore, we normalized the calculated SDFs to the range $[-1, 1]$ and employed a tanh-activated output layer in the regression head.

\subsection{Loss Function}
We adopted different loss functions for segmentation and regression heads.

\subsubsection{Segmentation Head}
In the head of predicting segmentation maps, we adopted binary cross-entropy loss and dice loss. The binary cross-entropy loss is very common in binary classification tasks, but the loss may miss part of the high-level information as it only considers the loss pixel by pixel. Therefore, we introduced the dice loss when training the network.

We set the probability of the $i$th point belonging to the whole brain as $\hat{p_i}$ and the ground truth as $y_i$, and the combined loss function of segmentation head $\mathcal{L}_{seg}$ can be defined as
\begin{align*}
\mathcal{L}_{seg} &= \mathcal{L}_{binary} + \mathcal{L}_{dice}\\
&= \sum_i (-y_i log(\hat{p_i}) - (1 - y_i)log(1 - \hat{p_i}))\\
&\quad + 1 - \frac{2\sum_i y_i \hat{p_i} + \epsilon}{\sum_i y_i + \sum_i \hat{p_i} + \epsilon}
\end{align*}
where $\epsilon$ is a small term introduced to avoid dividing by zero.

\begin{table*}[h]
\begin{center}
\begin{tabular}{c|c|c|c|c|c}
\hline
Model                                                                       & Volumetric Dice & Surface Dice & HD & HD95 & ASSD \\ \hline
a        & \textbf{0.9663$\pm$0.0514} & 0.9108$\pm$0.1503 &   66.6318$\pm$32.8307   & 2.9181$\pm$4.1790   &  0.5544$\pm$0.8412     \\ 
b        & 0.9529$\pm$0.0474 & 0.8534$\pm$0.1390 &   72.7802$\pm$26.5825  &  5.1333$\pm$3.7582   &  0.8929$\pm$0.7612    \\ 
c       & 0.9660$\pm$0.0509 & \textbf{0.9110$\pm$0.1501} &  58.9821$\pm$35.1389    & \textbf{2.0843$\pm$2.1847}   &  \textbf{0.4653$\pm$0.7870}    \\ 
d & 0.9653$\pm$0.0507 & 0.9087$\pm$0.1495   & \textbf{52.5453$\pm$26.4817}          &  2.2245$\pm$2.2473 &  0.5011$\pm$0.7876      \\ \hline
\end{tabular}
\caption{Quantitative results: mean and standard deviation of the volumetric dice, surface dice, HD, HD95, and ASSD. a) 2D U-Net with $\mathcal{L}_{seg}$, b) 3D U-Net with $\mathcal{L}_{seg}$, c) Our method with $\mathcal{L}_{seg} + \mathcal{L}_1$, d) Our method with $\mathcal{L}_{seg} + \mathcal{L}_1 + \mathcal{L}_{laplacian}$.}
\label{quantity}
\end{center}
\end{table*}

\subsubsection{Regression Head}
In the head of predicting regression maps, we adopted $\mathcal{L}_1$ loss as one part of the loss function of regression head $\mathcal{L}_{reg}$ since $\mathcal{L}_1$ loss is robust to outliers \cite{xue2020shape}. Furthermore, based on the properties that SDFs should be smooth and continuous everywhere, we introduced a new loss called Laplacian loss $\mathcal{L}_{laplacian}$ \cite{li2017laplacian}. 

Given the predicted SDF results $y_{pred}$ and the SDFs' ground truth $y_{true}$, Laplacian loss $\mathcal{L}_{laplacian}$ can be expressed as 
\begin{equation*}
\mathcal{L}_{laplacian} = \frac{1}{n}\sum_{i = 1}^n |\mathbf{D}_{xy}^2 * y_{true} - \mathbf{D}_{xy}^2 * y_{pred}|
\end{equation*}
where $\mathbf{D}_{xy}^2$ refers to the discrete Laplacian filter and $n$ refers to the number of pixels in each brain slice.
\begin{equation*}
\mathbf{D}_{xy}^2 = \begin{bmatrix}
0 & 1 & 0 \\
1 & -4 & 1 \\
0 & 1 & 0
\end{bmatrix}
\end{equation*}



\subsubsection{Final Loss}
When training the segmentation maps and regression distance maps simultaneously, the final loss is defined as
\begin{equation*}
\mathcal{L} = \mathcal{L}_{seg} + \mathcal{L}_{reg} = \mathcal{L}_{binary} + \mathcal{L}_{dice} + \mathcal{L}_1 + \mathcal{L}_{laplacian}
\end{equation*}

\subsection{Post Processing}
To generate the brain isosurface, we stacked the predicted segmentation maps for each patient and used marching cubes to complete the mesh reconstruction. We also tried to apply marching cubes to the SDFs but found that the brain isosurfaces are a little bit fuzzy, so we did not list the results in Section \ref{sec: experiments}.

\section{Experiments and Results}
\label{sec: experiments}

\subsection{Dataset and Experimental Details}
The dataset consisted of brain MRI slices from 111 patients. Medical doctors manually annotated the brain partitions in the slices, and we employed the results as the ground truth (gold standard).

We implemented four different models to validate our method. Details of the models are listed in Table \ref{quantity}.

We chose MRI slices from 66 and 22 patients for the training and validation phases (144 $\sim$ 190 slices per patient), respectively, reserving the remaining slices from 23 patients for the testing phase. All models were trained with the Adam optimizer with a decaying learning rate initialized at $1e-5$. The initial number of channels was 32, and the maximum number of channels after the last downsampling operator was 512. The input size of 2D U-Net was $512\times 512$, while that of 3D U-Net was $16\times 512\times 512$. The batch sizes of 2D U-Net and 3D U-Net were set as 8 and 1, separately, owing to memory limitation. All models were trained for 50 epochs, and we employed four-fold cross-validation to choose the best models of four experiment settings respectively (highest dice score in the validation set). Moreover, we performed all evaluations on the testing set to derive the quantitative results. All experiments were performed on an NVIDIA 1080TI GPU and the code was developed in Keras. 

\subsection{Evaluation Metrics}
We employed separate evaluation metrics for the whole-brain surface extraction in all experiments. Specifically, we used the volumetric dice score, surface dice score, HD, 95\% HD (HD95), and ASSD of each patient as the evaluation metrics. The volumetric dice score computes the dice score of the whole brain between the predicted mask and ground truth, while the surface dice score measures the overlap of two surfaces instead of two volumes. Furthermore, we set the tolerance of the surface to 1. HD, HD95, and ASSD were used to measure the difference between two different 3D representations of the same brain isosurface, which could be regarded as a shape-aware comparison.

\begin{figure*}[h]
\centering
\begin{subfigure}[t]{\dimexpr0.13\textwidth+20pt\relax}
    \makebox[20pt]{\raisebox{25pt}{\rotatebox[origin=c]{0}{$a)$}}}%
    \includegraphics[width=\dimexpr\linewidth-20pt\relax]
    {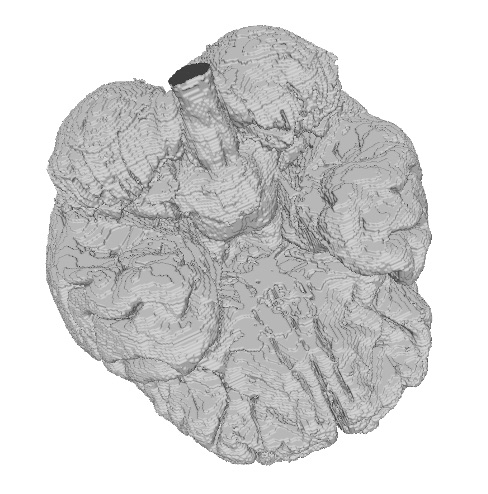}
    \makebox[20pt]{\raisebox{25pt}{\rotatebox[origin=c]{0}{$b)$}}}%
    \includegraphics[width=\dimexpr\linewidth-20pt\relax]
    {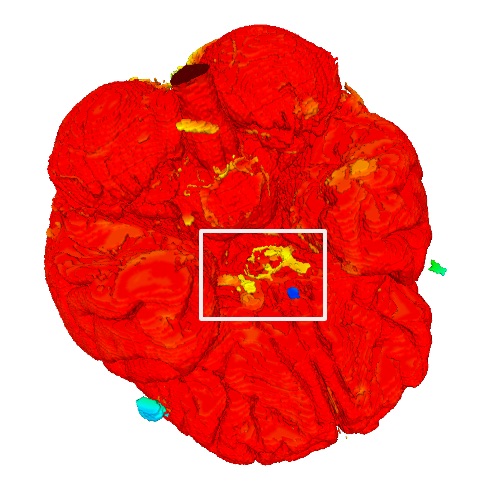}
    \makebox[20pt]{\raisebox{25pt}{\rotatebox[origin=c]{0}{$c)$}}}%
    \includegraphics[width=\dimexpr\linewidth-20pt\relax]
    {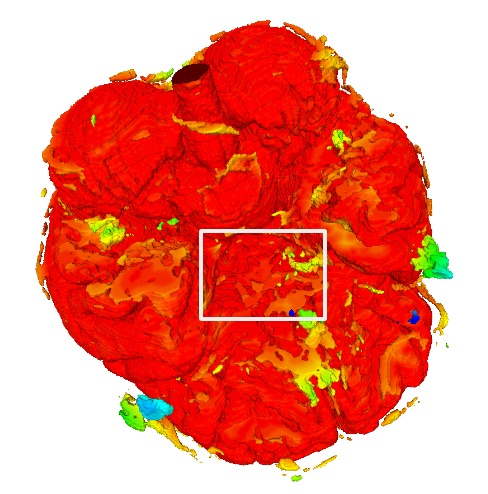}
    \makebox[20pt]{\raisebox{25pt}{\rotatebox[origin=c]{0}{$d)$}}}%
    \includegraphics[width=\dimexpr\linewidth-20pt\relax]
    {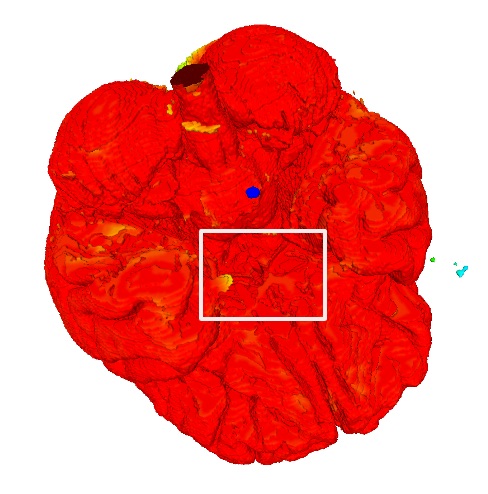}
    \makebox[20pt]{\raisebox{25pt}{\rotatebox[origin=c]{0}{$e)$}}}%
    \includegraphics[width=\dimexpr\linewidth-20pt\relax]
    {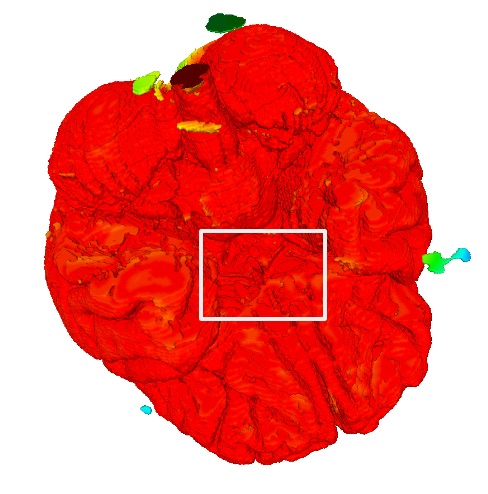}
\end{subfigure}\hfill
\begin{subfigure}[t]{0.13\textwidth}
    \includegraphics[width=\textwidth]  
    {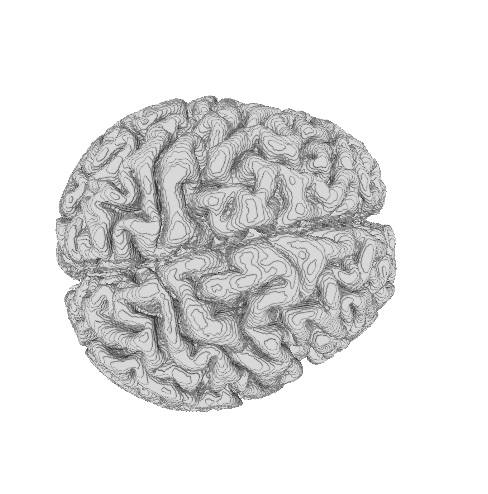}
    \includegraphics[width=\textwidth]  
    {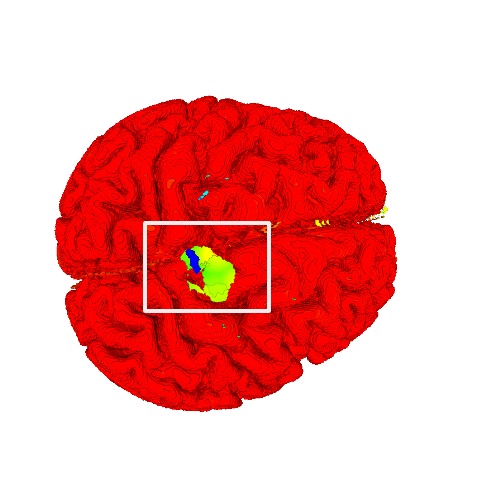}
    \includegraphics[width=\textwidth]  
    {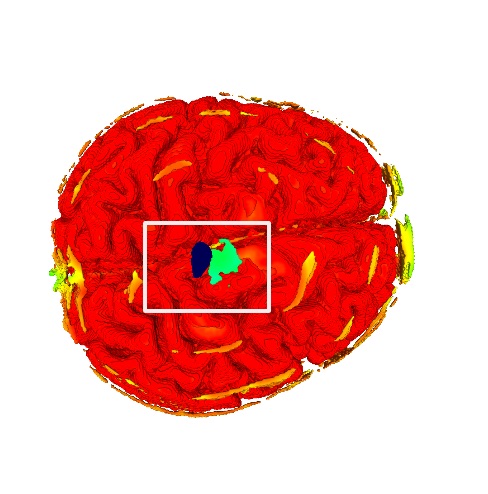}
    \includegraphics[width=\textwidth]  
    {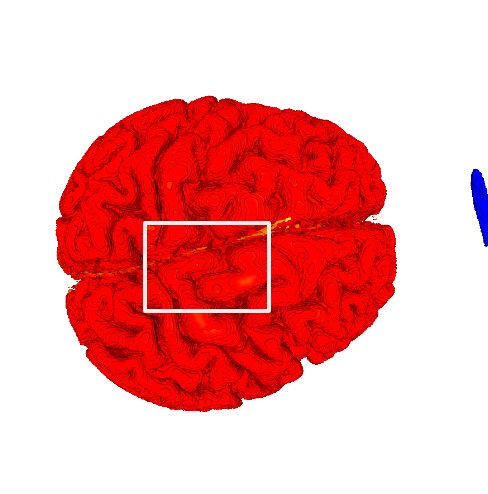}
    \includegraphics[width=\textwidth]  
    {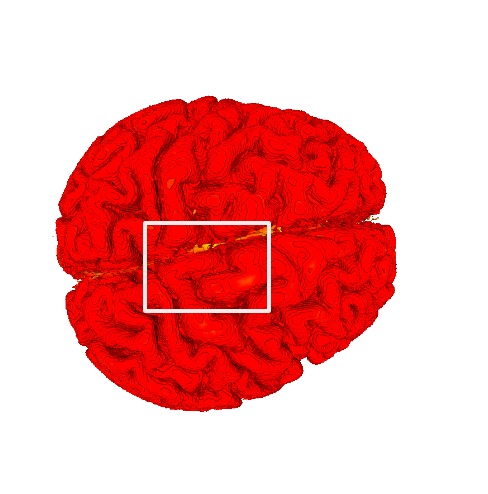}
\end{subfigure}\hfill
\begin{subfigure}[t]{0.13\textwidth}
    \includegraphics[width=\textwidth]  
    {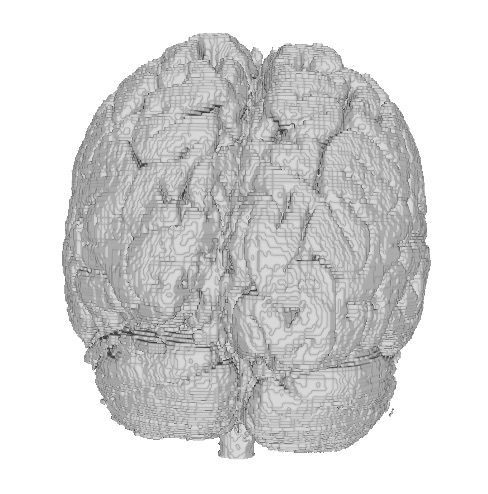}
    \includegraphics[width=\textwidth]  
    {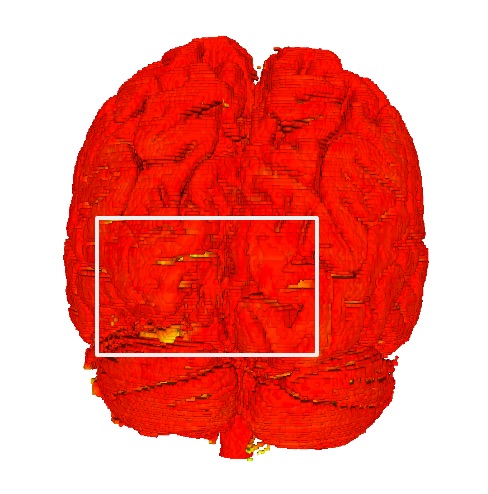}
    \includegraphics[width=\textwidth]  
    {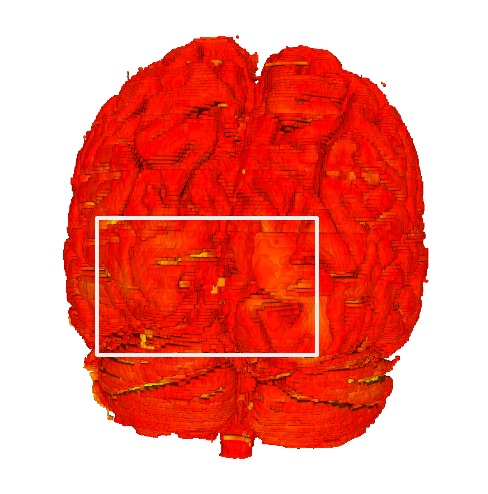}
    \includegraphics[width=\textwidth]  
    {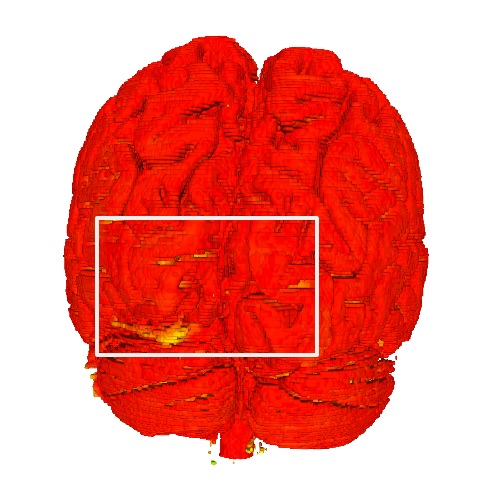}
    \includegraphics[width=\textwidth]  
    {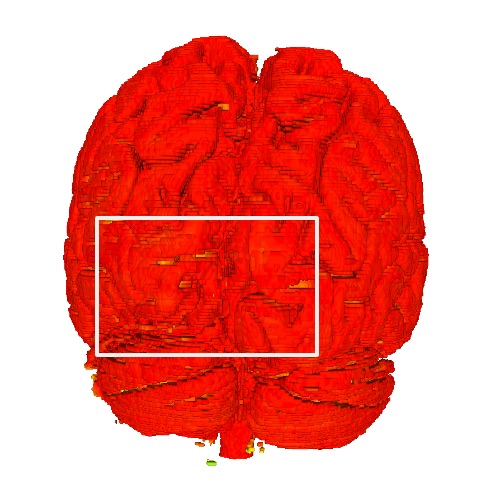}
\end{subfigure}\hfill
\begin{subfigure}[t]{0.13\textwidth}
    \includegraphics[width=\textwidth]  
    {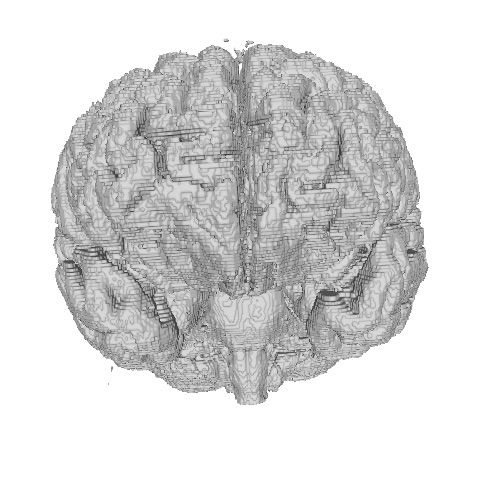}
    \includegraphics[width=\textwidth]  
    {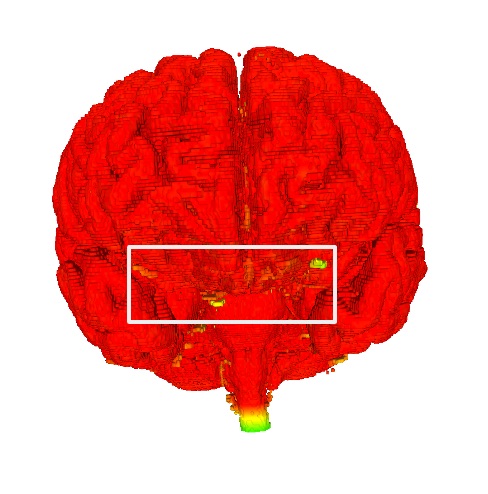}
    \includegraphics[width=\textwidth]  
    {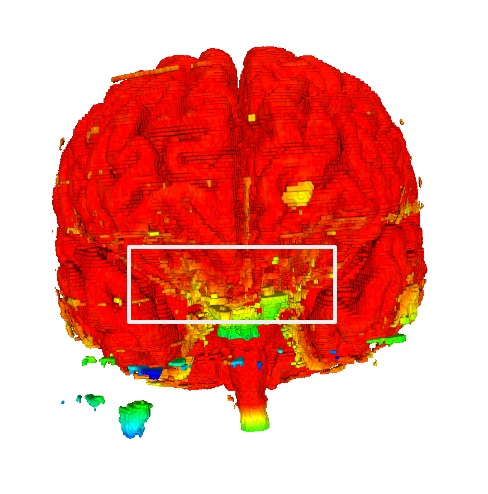}
    \includegraphics[width=\textwidth]  
    {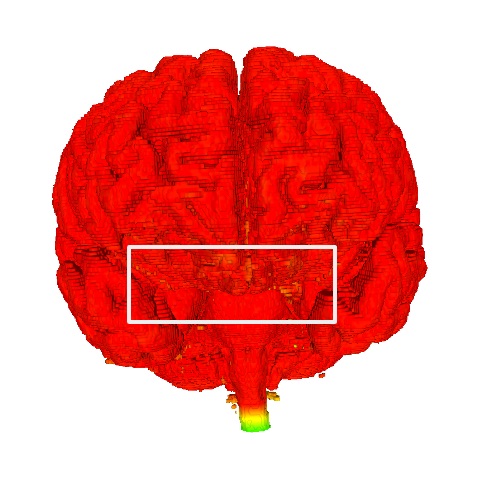}
    \includegraphics[width=\textwidth]  
    {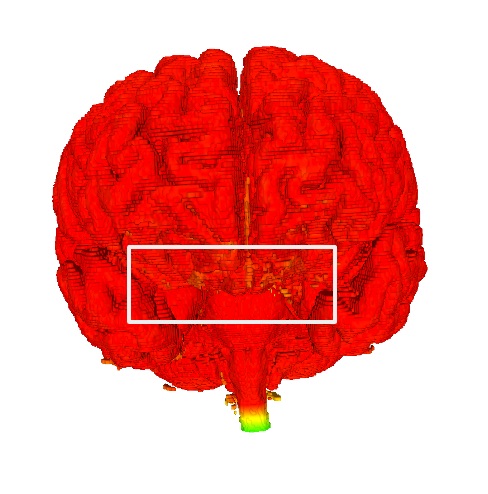}
\end{subfigure}\hfill
\begin{subfigure}[t]{0.13\textwidth}
    \includegraphics[width=\textwidth]  
    {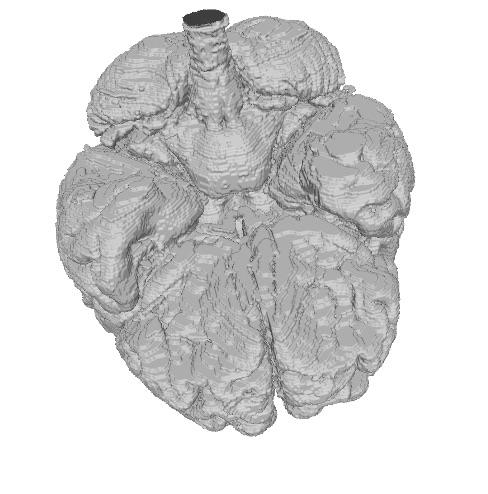}
    \includegraphics[width=\textwidth]  
    {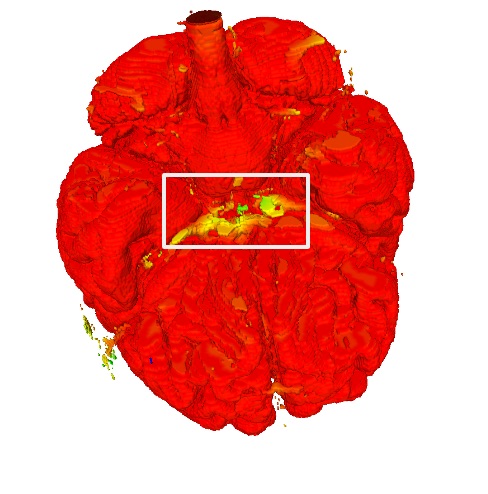}
    \includegraphics[width=\textwidth]  
    {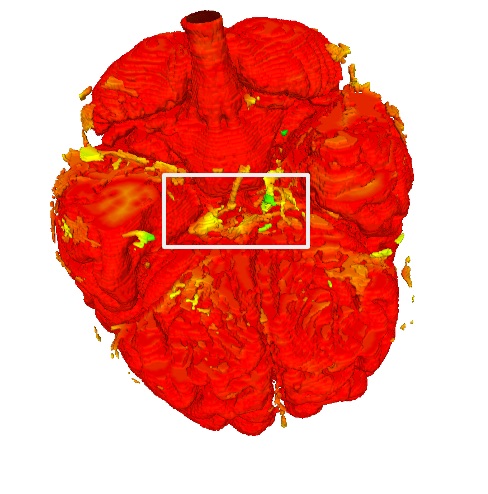}
    \includegraphics[width=\textwidth]  
    {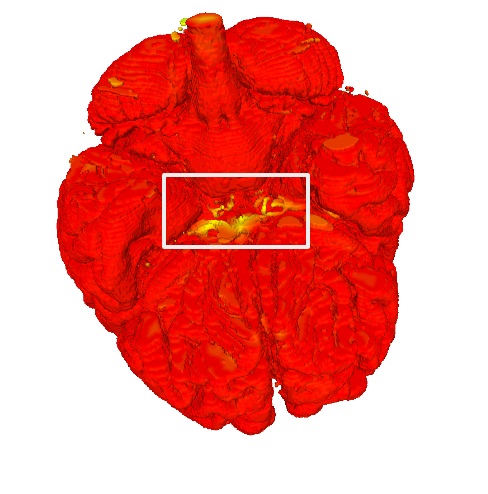}
    \includegraphics[width=\textwidth]  
    {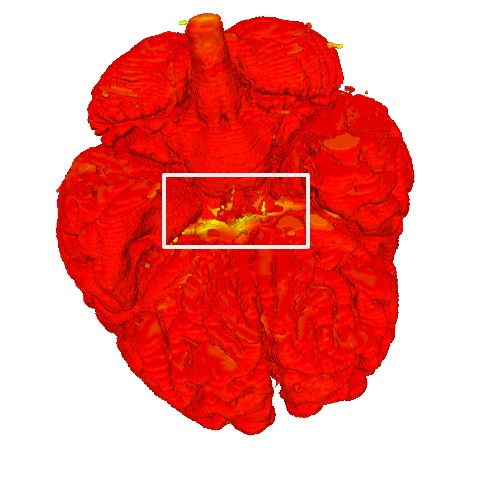}
\end{subfigure}\hfill
\begin{subfigure}[t]{0.13\textwidth}
    \includegraphics[width=\textwidth]  
    {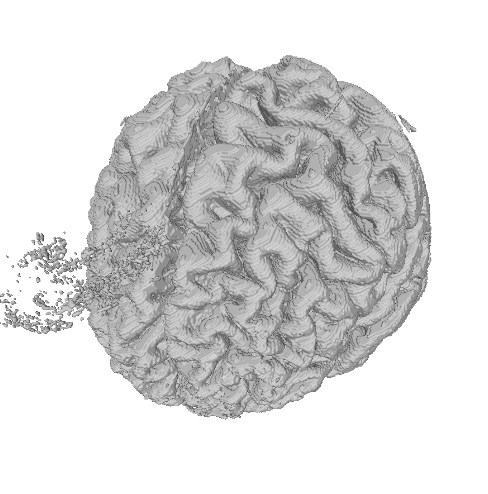}
    \includegraphics[width=\textwidth]  
    {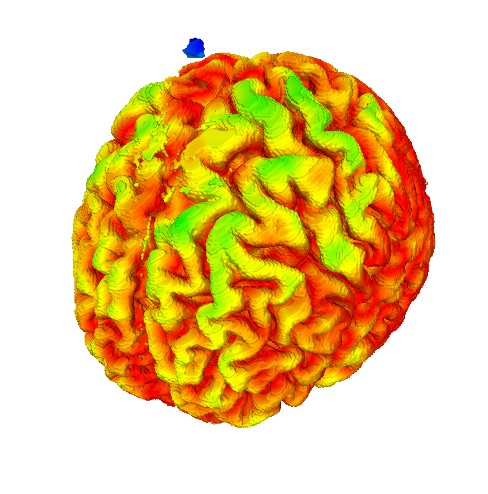}
    \includegraphics[width=\textwidth]  
    {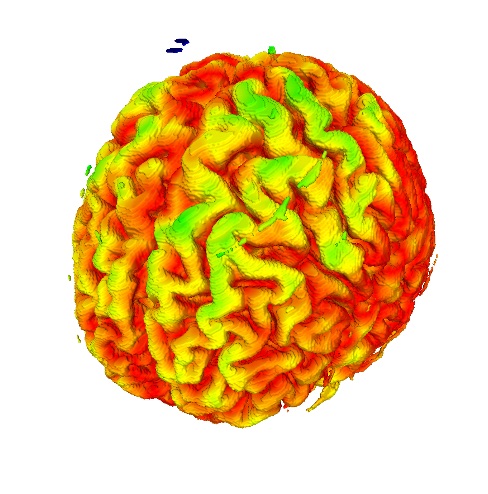}
    \includegraphics[width=\textwidth]  
    {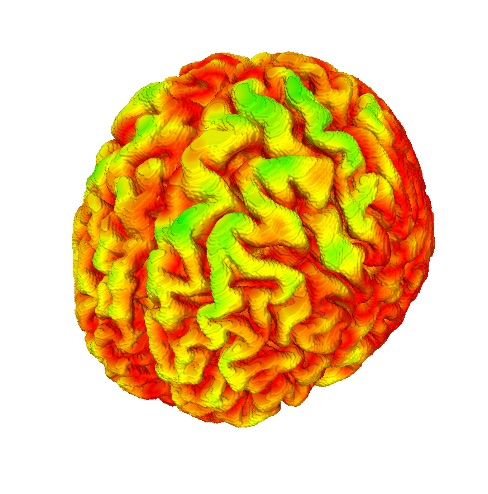}
    \includegraphics[width=\textwidth]  
    {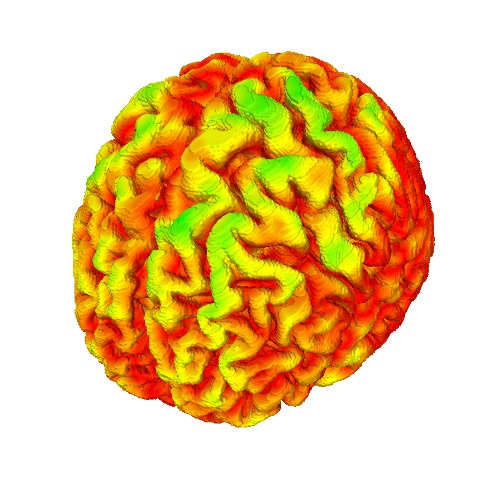}
\end{subfigure}\hfill
\includegraphics[width=\textwidth]{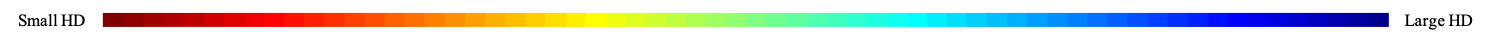}
\caption{Qualitative results: Each image shows the brain isosurface in 3D view. a) ground truth, b) prediction from 2D U-Net baseline, c) prediction from 3D U-Net baseline, d) prediction from our method without Laplacian loss, e) prediction from our method. HDs are visualized in red-blue colormap where red refers to small HD and blue refers to large HD. Special regions are highlighted with a white box to show where our method improves the quality of brain isosurfaces.}
\label{quality}
\end{figure*}

\subsection{Quantitative Results}

We used 2D U-Net and 3D U-Net as the baseline of our method. Table \ref{quantity} gives a quantitative comparison of the results on our dataset. We observe from Table \ref{quantity} that training a U-Net-based convolution neural network with an extra head learning from SDFs can achieve comparable results in the metrics of volumetric and surface dice scores. We assume that the whole brain is a large organ and it is not difficult to predict, even at baseline. However, 3D U-Net does not perform well in the brain isosurface extraction task. We assume that this is because 3D U-Net has more training hyperparameters and our dataset is not large enough to obtain a well-trained network. Moreover, after applying four downsampling operations, the intermediate feature maps would easily lose most or even all information of the first dimension.

In addition, we compared our proposed methods with the baseline networks (2D and 3D U-Net) on the shape-aware evaluation metrics (HD, HD95, and ASSD). The comparison demonstrates that our method outperforms HD, HD95, and ASSD for these metrics. The results prove that adding a head to the baseline network can be beneficial for learning shape information and that our introduced Laplacian loss can reduce HD at the 3D level, which is suitable for mesh reconstruction of the organs. 

\subsection{Qualitative Results}
In Fig. \ref{quality}, we show brain isosurface results of six patients obtained from 2D U-Net, 3D U-Net, and our method. We computed HD between the predicted brain isosurface and ground truth, and visualized HD in red-blue colormap (red - small HD, blue - large HD). We highlight special regions with a white box to show where our method improves the quality of brain isosurfaces. We can clearly see that adding one extra head to learn the information of SDFs can help produce a clearer and more stable brain isosurface than the baseline networks and that our proposed Laplacian loss can reduce HD at the 3D level (HD is sensitive to outliers). Moreover, since the ground truth is annotated manually in each slice separately, it may contain some isolated false positives in 3D view, and the last column shows that our method can still achieve a good brain isosurface even in this case.

\section{Discussion and Conclusion}
\label{sec: discussion}
To address the problem of brain isosurface reconstruction, we propose a shape-aware segmentation network that incorporates the information of SDFs. We observed that 2D U-Net performs better than 3D U-Net in the segmentation task of large organs such as the brain. Furthermore, we proved that our method can achieve volumetric and surface dice scores comparable with those of 2D U-Net. Moreover, our method can decrease the HD and ASSD when conducting mesh reconstruction from the predicted segmentation maps.

However, we are not taking full advantage of SDFs, which have been predicted by the regression head. Our future work will be to evaluate the predicted SDFs and perform the reconstruction directly from the predicted SDFs.

\clearpage
\section{Compliance with Ethical Standards}
\label{sec:ethics}
All procedures involving human participants were in accordance with the ethical standards of the institutional research committee and with the 1964 Helsinki Declaration and its later amendments. Informed consent was obtained from all participants included in the study. This study was approved by the Research Ethics Committee of the University of Tokyo.

\section{Acknowledgments}
\label{sec:acknowledgments}
This work was supported by JST CREST Grant Number JPMJCR17A1 and AMED under Grant Number JP18he1602001, Japan. The authors would like to thank Zheyuan Cai, Ding Xia and Yang Zhou for their helpful comments.

\bibliographystyle{IEEEbib}
\bibliography{strings,refs}

\end{document}